\RequirePackage{fix-cm}
\documentclass[smallextended]{svjour3}       
\smartqed  
\usepackage{graphicx}
\usepackage{amsmath}
\usepackage{adjustbox}
\usepackage{subfig}
\usepackage{cases}
\usepackage{algorithm}
\usepackage{algpseudocode}
\begin{document}

\title{OCmst: One-class Novelty Detection using Convolutional Neural Network and Minimum Spanning Trees
}

\titlerunning{OCmst}        

\author{Riccardo La Grassa \and Ignazio Gallo \and Nicola Landro}

\authorrunning{La Grassa, R et al} 

\institute{La Grassa Riccardo, Gallo Ignazio, Landro Nicola \\
              University of Insubria \\
              \email{rlagrassa, ignazio.gallo, nlandro \{@uninsubria.it\}}
}

\date{Received: date / Accepted: date}

\maketitle

\begin{abstract}
We present a novel model called One Class Minimum Spanning Tree (OCmst) for novelty detection problem that uses a Convolutional Neural Network (CNN) as deep feature extractor and graph-based model based on Minimum Spanning Tree (MST).
In a novelty detection scenario, the training data is no polluted by outliers (abnormal class) and the goal is to recognize if a test instance belongs to the normal class or to the abnormal class.
Our approach uses the deep features from CNN to feed a pair of MSTs built starting from each test instance. 
To cut down the computational time we use a parameter $\gamma$ to specify the size of the MST's starting to the neighbours from the test instance.
To prove the effectiveness of the proposed approach we conducted experiments on two publicly available datasets, well-known in literature and we achieved the state-of-the-art results on CIFAR10 dataset.
\end{abstract}

\vspace{0.5cm}
\vspace{0.1cm}
\textit{This paper is under review to "Neural Computing and Applications".}

\section{Introduction}
\label{intro}
One-class novelty detection refers to the recognize of abnormal patters on data recognized as normal.
Abnormal data, also know as outliers, anomaly or alien data are patters who belong to the different classes than normal class.
The goal of the novelty detection field is to distinguish anomaly patterns which are different by normal and classify them.
The capability of many machine learning technique, in the field of novelty and outlier detection is to decide whether a new instance belongs to the same distribution or if it has different behaviour such as to be considered as outlier. Typically, outliers detection is unsupervised and the goal is to recognize the density of clusters to discover possible outliers. 
We focus on novelty detection with semi-supervised approach trained without outliers, whose goal is to decide whether a new observation is an outlier. 
In literature, many conventional one-class classifiers that resolve the novelty detection problem exists, such OCSVM \cite{scholkopf2001estimating}, MST\_CD \cite{juszczak2009minimum,GrassaGCO19,DBLP_LAGRASSA}.
Due to the high complexity of some data types, such as images and audio signals, these novelty detection methods suffer also of bad performance on high-dimensional data and then dimensionality reduction techniques are required.
To solve this problem techniques as Principal component analysis (PCA) and singular value decomposition (SVD) are commonly used to dimensionality reduction or classical feature selection using statistical metrics. 
These approaches are task-dependent and they need to an expert supervisor.
In contrast to the traditional machine learning approach, deep learning models such as GAN  \cite{schlegl2017unsupervised,perera2019ocgan} and Deep One-Class (DOC) \cite{perera2019learning}, are able to extract these features independently from the particular task to be solved. 
In literature, few deep learning approaches exist to solve novelty detection problems.
Our focus, in this paper, is to investigate a generic method for one-class classification using a convolutional neural network as deep feature extractor and minimum spanning trees, able to pattern recognition.
To the best of our knowledge, none of the previous work used convolutional neural network jointly with a graph-based model. 
In this work, we extend two our graph-based previous works \cite{GrassaGCO19,DBLP_LAGRASSA} and we use them as one-class novelty detection approach exploiting deep features.
Our work makes the following contributions:
\begin{enumerate}
    \item We extend our previous works on MST and use jointly with a convolutional neural network to solve novelty detection problems.
    \item To prove the effectiveness and robustness of the proposed approach we evaluate on two well-known available datasets where we achieve the state-of-the-art across many tasks.
\end{enumerate}

\section{Related work}
In this section, we briefly introduce the main approaches used for novelty detection and highlight advantages and disadvantages.
In general, the problems of One-class classification is harder than the problem of normal two-class classification. For normal classification, the decision boundary is supported from both sides by examples of each the classes. In One-class classification only one side of the boundary is covered and available and it is hard to find the best separation of the target and the outliers class.
Anomaly detection and one-class classification are problems related to one-class novelty detection \cite{chalapathy2019deep}. Both have similar objectives – to detect out-of-class samples given a set of in-class samples. A hard label is expected to be assigned to a given image in one-class classification; therefore its performance is measured using detection accuracy and F1 score. In contrast, novelty detection is only expected to associate a novelty score to a given image; therefore the performance of novelty detection is measured using a Receiver Operating Characteristic (ROC) curve.
The supervised approach offers a better approach in terms of performance than unsupervised novelty detection techniques \cite{gornitz2013toward}.
Models that use this approach learns the hyperplane of separation or a generic decision boundary to discriminate data instances and then, to predict whether test instances belong to this boundary of if it lies outside.
Deep model based on a supervised approach fails when the features space is highly and non-linear and these methods require various data from the training of both classes (normal and abnormal) that usually are unavailables.
Against, the unsupervised approach is used to distinguish normal and abnormal class without know labels data instances. Usually, these methods are used to automate the process of data labelling. Autoencoder is used as unsupervised deep architecture in novelty detection \cite{baldi2012autoencoders}\cite{hinton2006reducing}. In the case with unavailable data labelled, this approach offers good results but, often it is a challenge to learn common features among instances in high dimensionality and with a high non-linear distribution of data.
Semi-supervised in novelty detection are widely used and they assume that all training instances of only a class is known and the goal is to recognize is an object is predicted as normal or abnormal, for instance OCSVM, SVDD and others.
The main idea for the one-class support vector method (OCSVM) is to separates all the data from feature space F and maximize the distance from a hyperplane to the origin. In contrast with traditional SVM, OCSVM learns a decision boundary that achieves maximum separation between the samples of the known class and the origin. 
A binary discriminative function is used to assign a positive label if the test belongs to a region or negative whether it lies out of the boundary.
Instead, to consider a hyperplane, SVDD \cite{Tax2004} takes a spherical boundary around the training data. The goal is to minimize the volume of this boundary such that a possible outlier lies outside.
The OCSVM and SVDD are closely related. Both can be adopted as novelty detection methods by considering distance to the decision boundary as the novelty score.
SVDD gives a higher owners correctness ration (true positive) in the case which a large variation in density exist among the normal-class instances. In such case, it starts to reject the low-density target points as outliers. Further, in the case, the data distribution is highly non-linear the probability to make the wrong prediction is high because is not possible to track a more detailed decision boundary around training data.
SVM is affected by the same problem and it does not perform very well when the data are overlapping, furthermore is not suitable for large datasets.
With the wide diffusion of deep learning, nowadays we can recognize a new type of models known as Hybrid models, able to solve novelty detection problems.
Deep learning models are used as deep features extractor and they are used as input to the traditional algorithms well-known in machine learning like one-class support vector machine, autoencoder+ocsvm \cite{andrews2016detecting}, autoencoder+knn\cite{song2017hybrid}, autoencoder+svdd \cite{kim2015deep}.
The main advantage of this hybrid technique is to reduce the curse of dimensionality and increase the discriminative power of features using neural networks.
A recently proposed approach is One-class neural networks (OCNN)\cite{chalapathy2018anomaly}\cite{ruff2018deep}\cite{perera2019learning} that combines a deep neural network while optimizing the data-enclosing boundary in output space.
Against the hybrid models, they do not require data for the classification and they outperform in terms of speed.
Intuitively, a disadvantage is the computational time required for training step and for model updates in a high dimensional input data.
Another technique to approach the novelty detection problems with neural networks is the 
GAN \cite{goodfellow2014generative}.
The generative adversarial network \cite{goodfellow2014generative} use a discriminator to distinguish between generated and real data simultaneously: when the discriminator can understand if the input was generated, the back-propagation update only the generator weight elsewhere it is updated only the discriminator weight.
The discriminator can be used as anomaly detector because it gets as results two class that the first represents the elements that are part of the class instead of the other class represent the element that is not in the class. Some examples of GAN used as anomaly detection is the AnoGAN \cite{schlegl2017unsupervised} or OCGan \cite{perera2019ocgan}.
It is possible to use the neural network for novelty detection with another technique: Autoencoder. The autoencoder can create a compressed version of the input and after it can generate again the input using this representation. It is possible to evaluate how well the decoded information is similar to the input information, so we can set a threshold \cite{marchi2015novel}, if the evaluation is bigger than the threshold the input is classified as new. Using variational autoencoder \cite{kingma2013auto} it is possible to improve the evaluation because it used to get as input the varied input, so only if some input is very different from the seen example it perform less.

\begin{figure*}
\centering
\includegraphics[width=\textwidth]{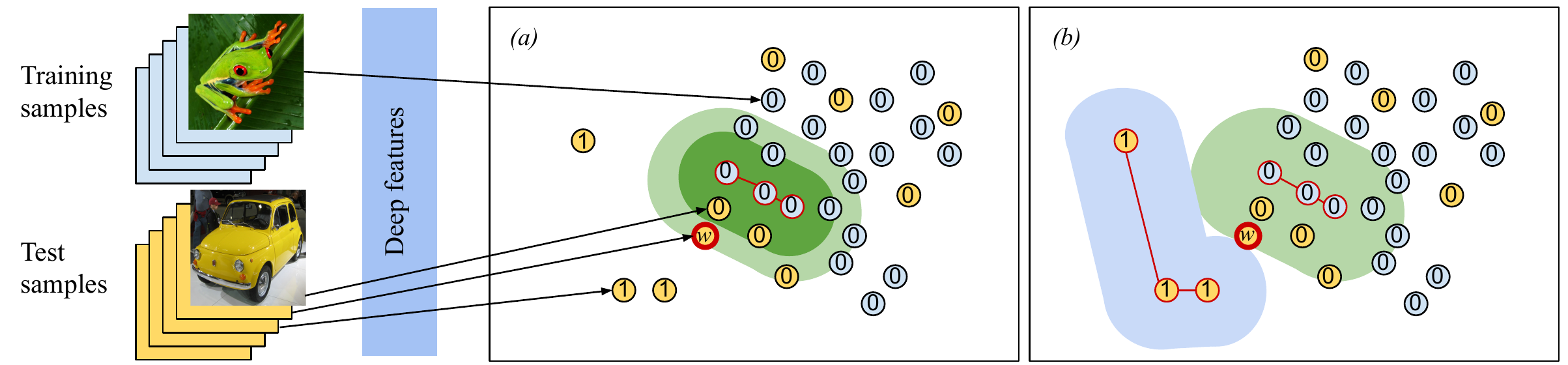}
    \caption{Overview of proposed novelty detection model.
    In the first step (a), features from training set and test set are extracted using a CNN.
    Labels 0, 1 or $w$ for test samples are assigned using a single MST with two different boundaries.
    In (b), for each test sample labelled as $w$ (uncertain value), two MST are created to assign a new label (0 or 1).
    }\label{fig:main-idea}
\end{figure*}

\section{Proposed Method}
In this paper, we propose a novel Deep Hybrid Model (DHM), called OCmst, that effectively explores the convolutional features for one class novelty detection in image classification.
Our goal is to label images, never seen during training, as belonging to the single class analyzed (0) or as anomalies (1). As graphically represented in Figure~\ref{fig:main-idea}, our goal is achieved in two main steps: \textit{(a)} through the use of a single MST to assign labels 0, 1 or $w$; \textit{(b)} through the use of two MSTs to resolve all previous $w$ labeled samples.

We use a generic convolutional neural network as feature extractors using only samples having a single class label, according to one-class classifiers. 
These train data are images transformed into deep features by a VGG19, pre-trained on Imagenet. 
The deep features extracted from a pre-trained CNN are used directly without any transformation and fed to our proposed OCmst method.

\begin{figure}
    \centering
        \subfloat[Accepted instance in green point]{{\includegraphics[width=9cm]{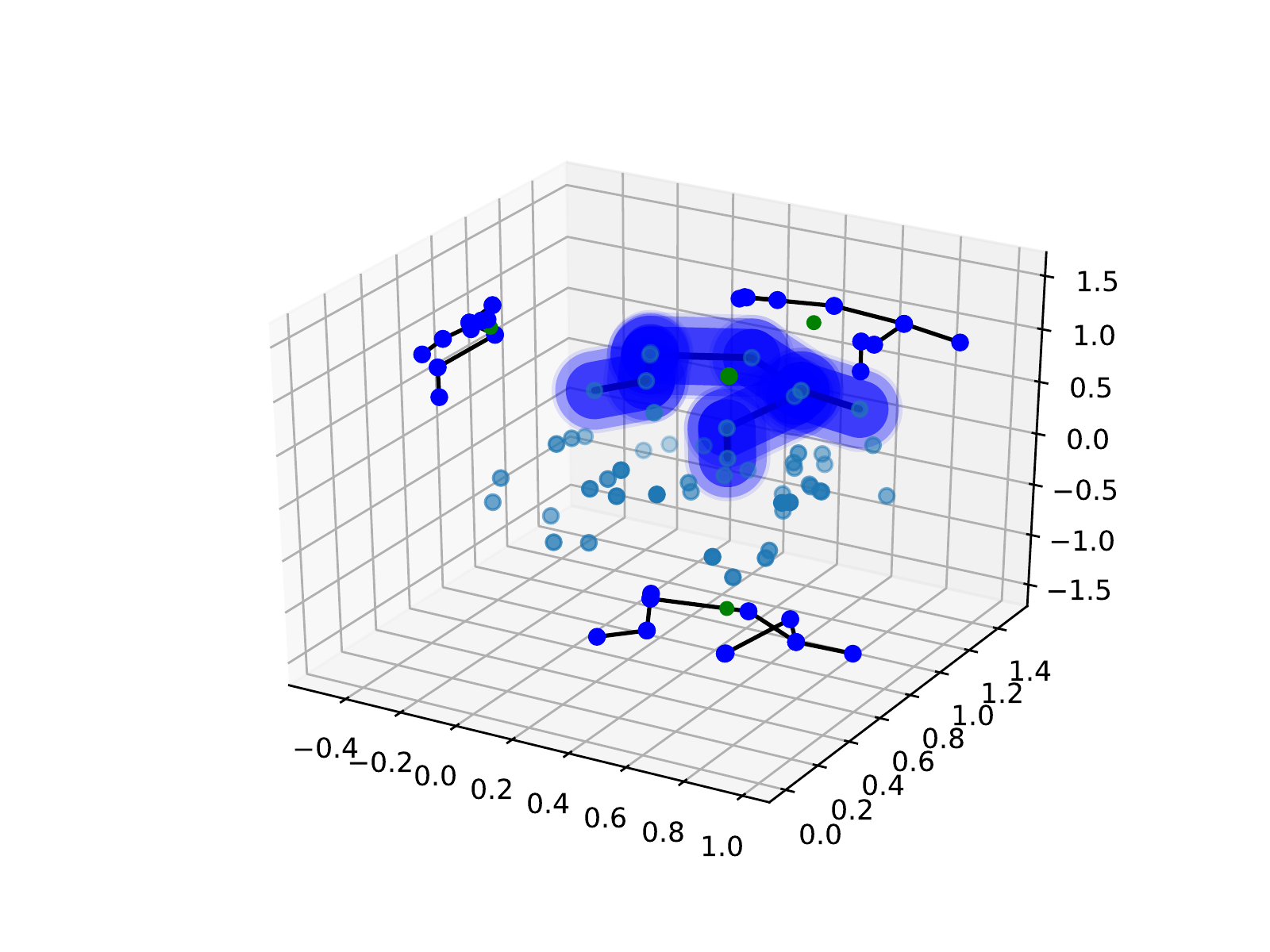}}}
    \\
        \subfloat[Rejected instance in red point]{{\includegraphics[width=9cm]{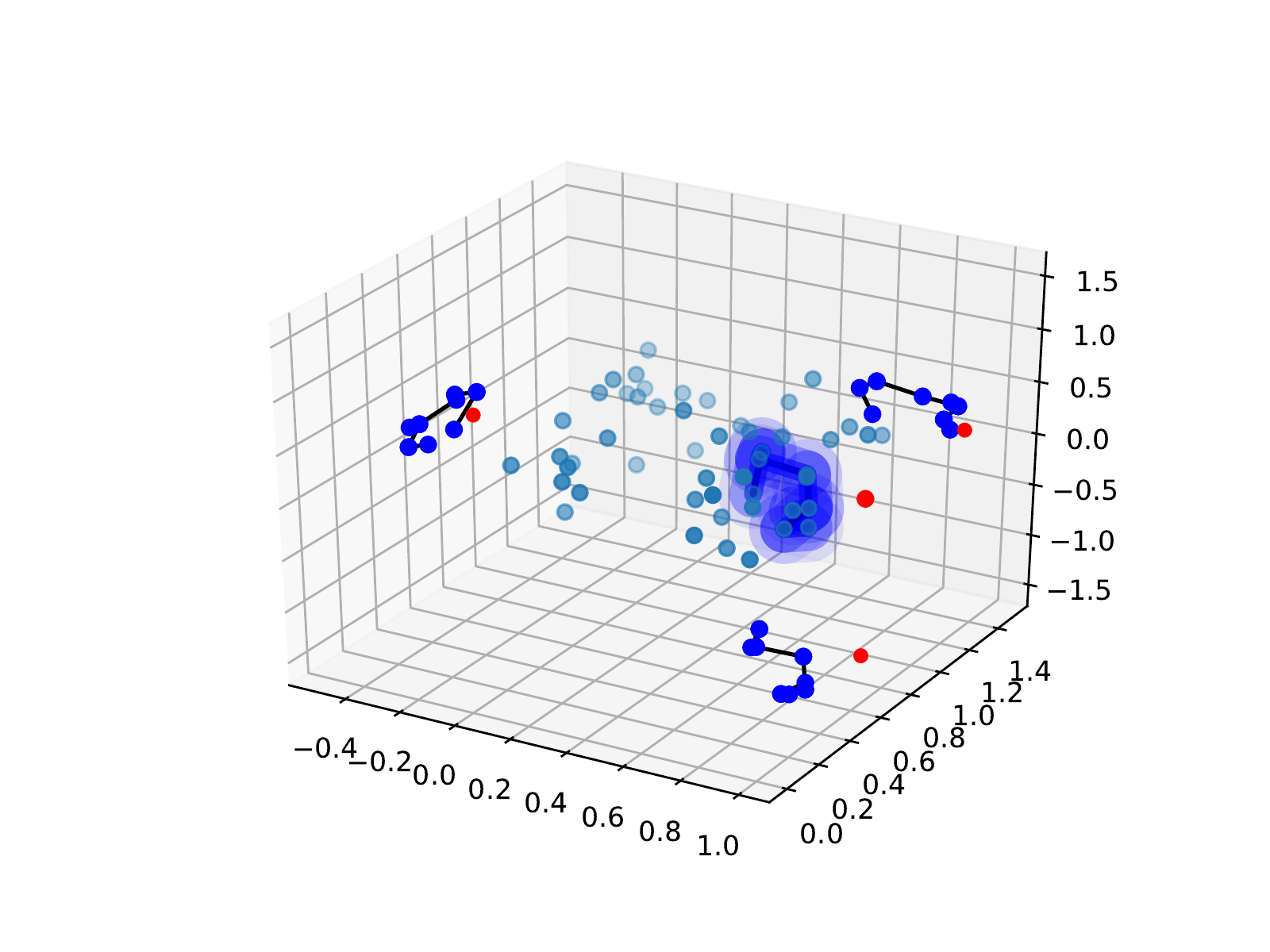}}}
    \\
    \subfloat[Uncertain instance in orange point]{{\includegraphics[width=9cm]{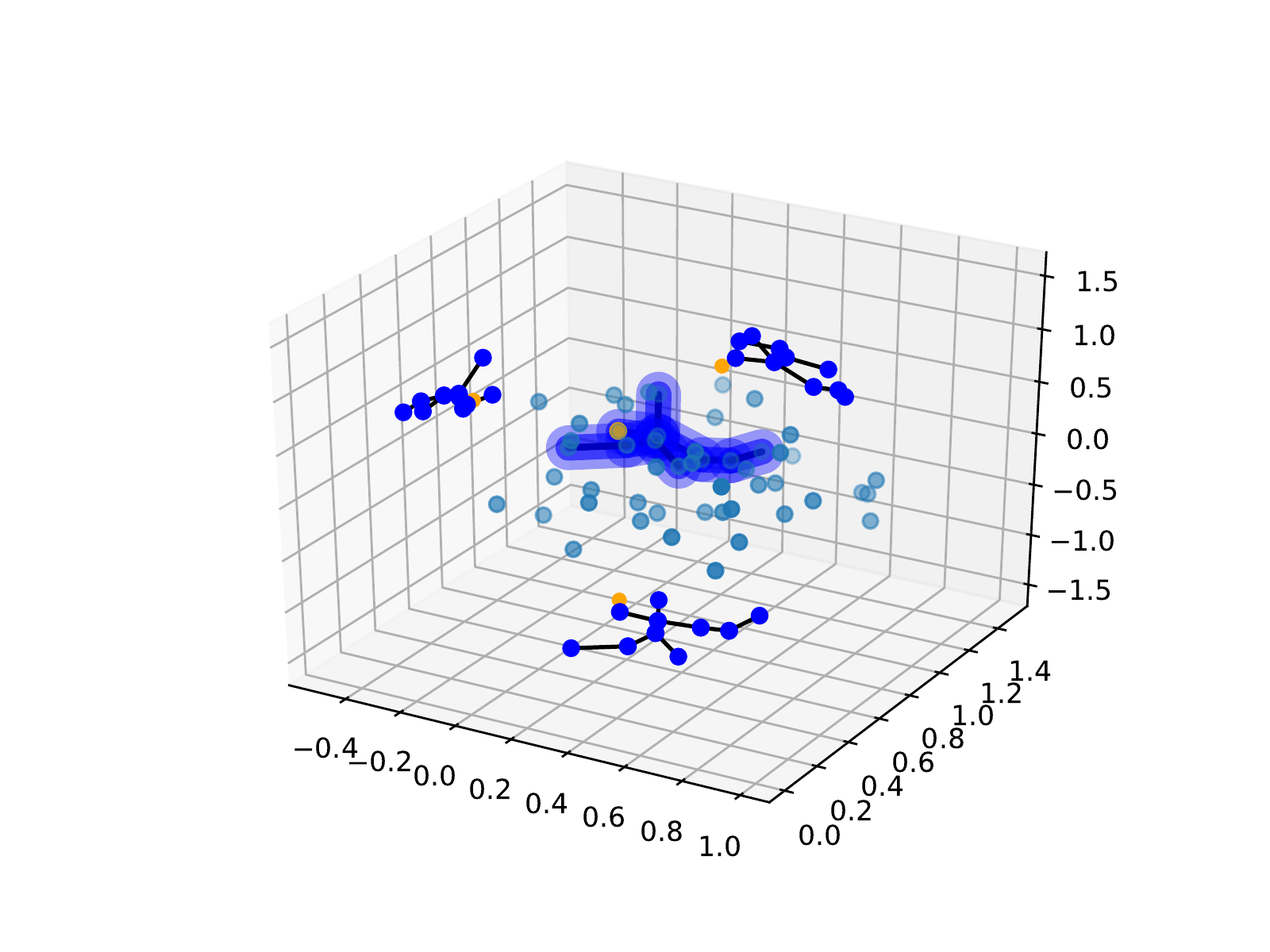}}}
    \caption{In this three figures, we show a 3D toy example of the first step in which the minimum spanning tree accept (a) or reject (b) the object. All instances strongly rejected (b) will be used as abnormal class to predict all remain instances recognized by "uncertain" (c) through a pair of MST. To better understand, we plot the projections on X,Y,Z axis in all three plots.}
    \label{fig:onemst}
    \label{fig:example}
\end{figure}

\begin{figure}
\centering
\includegraphics[width=\textwidth]{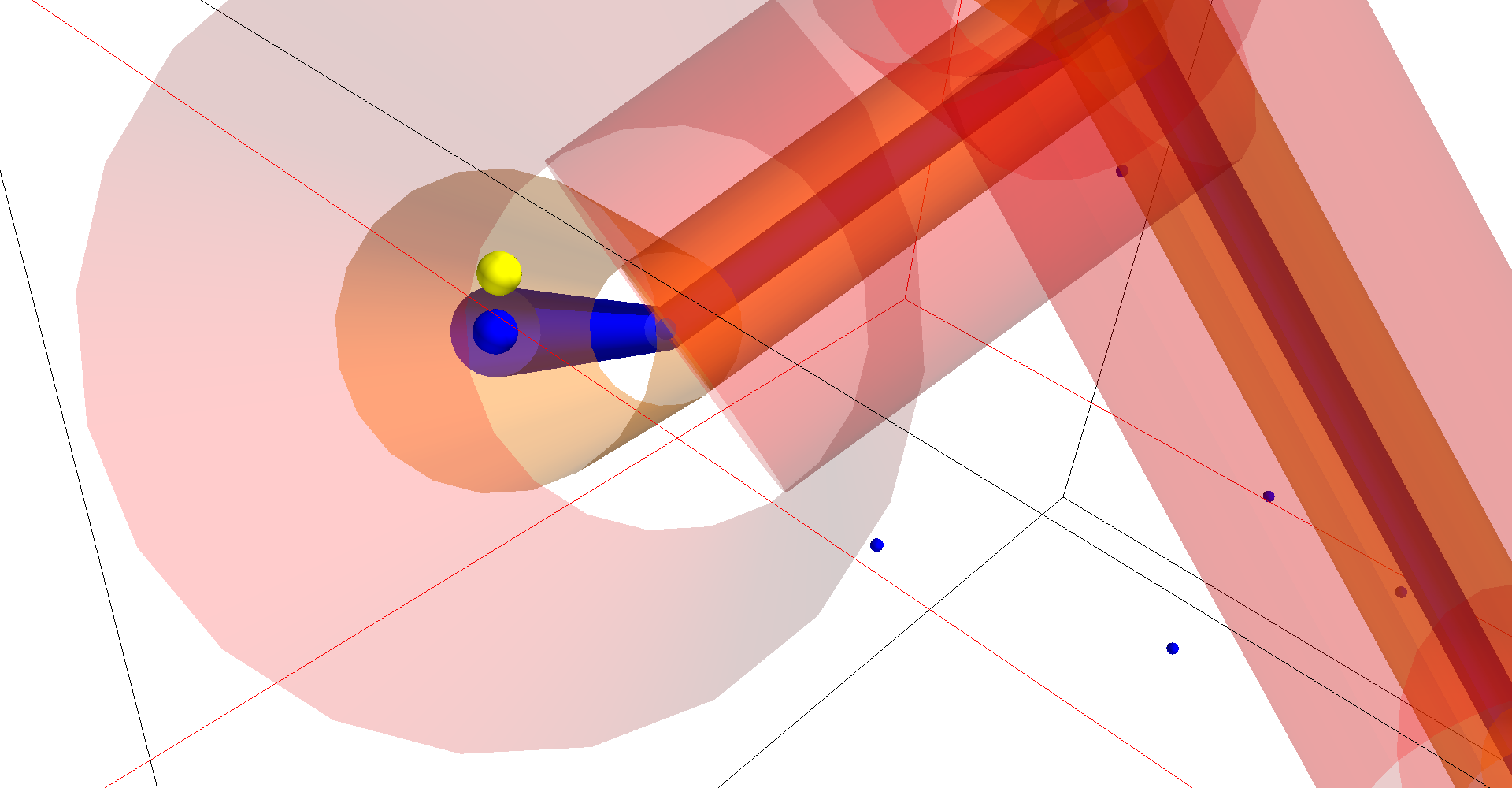}
    \caption{An example of uncertain instance. The yellow sample lies on the second boundaries decision and it will be classify as normal or abnormal sample in the next step.}
    \label{fig:uncertain}
\end{figure}

Going into detail, in our approach we can distinguish two main steps as part of the proposed OCmst model.
In the first step, when a new test sample $x$ is given, we select the first $\gamma$ instances $v \in G_{0}$ of the training set and create a ``complete graph'' using euclidean distance as the weight for each couple of edges.
All the training samples $v$ belong to the same normal class.
The selected $\gamma$ training samples of the normal class (only known class) are the closest to the sample $x$. 
Subsequently, we use the Kruskal's algorithm to find the minimum spanning tree using the previous selected $\gamma$ instances.
In contrast with previous work~\cite{GrassaGCO19}, we use two different boundaries around each MST to create the decision boundary and establish whether a test lies inside the first, second or out of the boundaries created.
In Figs.~\ref{fig:onemst} and~\ref{fig:uncertain}, we show on a 3-dimensional space a real case using OCmst on a toy dataset created to highlight three possible scenery (accepted/rejected/uncertain instance).
If $x$ lies in the second boundary, we label it as uncertain test, otherwise, if the sample lies in the first boundary then we label it as a normal class otherwise abnormal class label is assigned.

\begin{figure}
    \centering
        \subfloat[Accepted instance as normal class by first mst in blue triangle]{{\includegraphics[width=6cm]{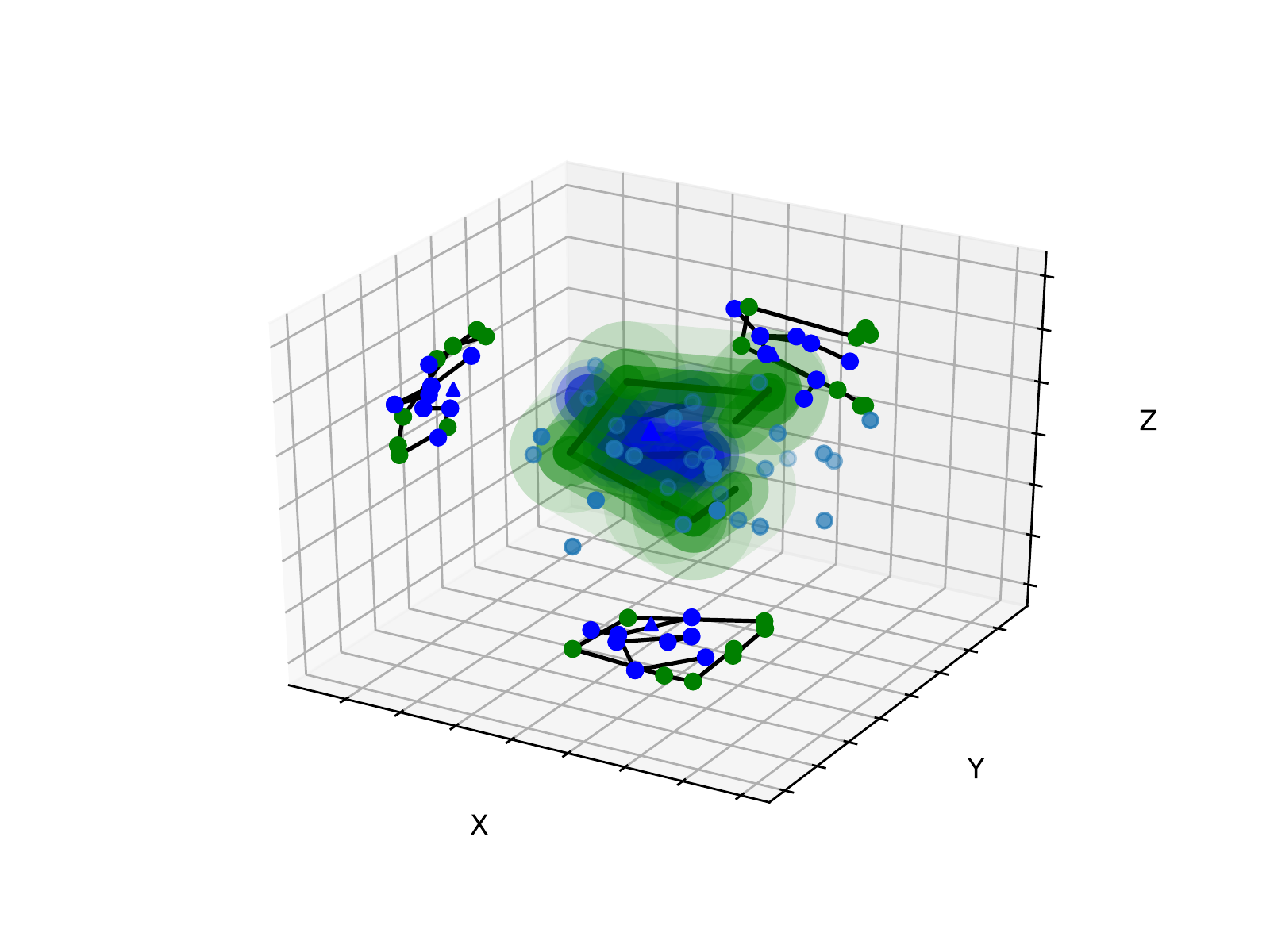}}}
        \subfloat[Accepted instance as abnormal class by second mst in green triangle]{{\includegraphics[width=6cm]{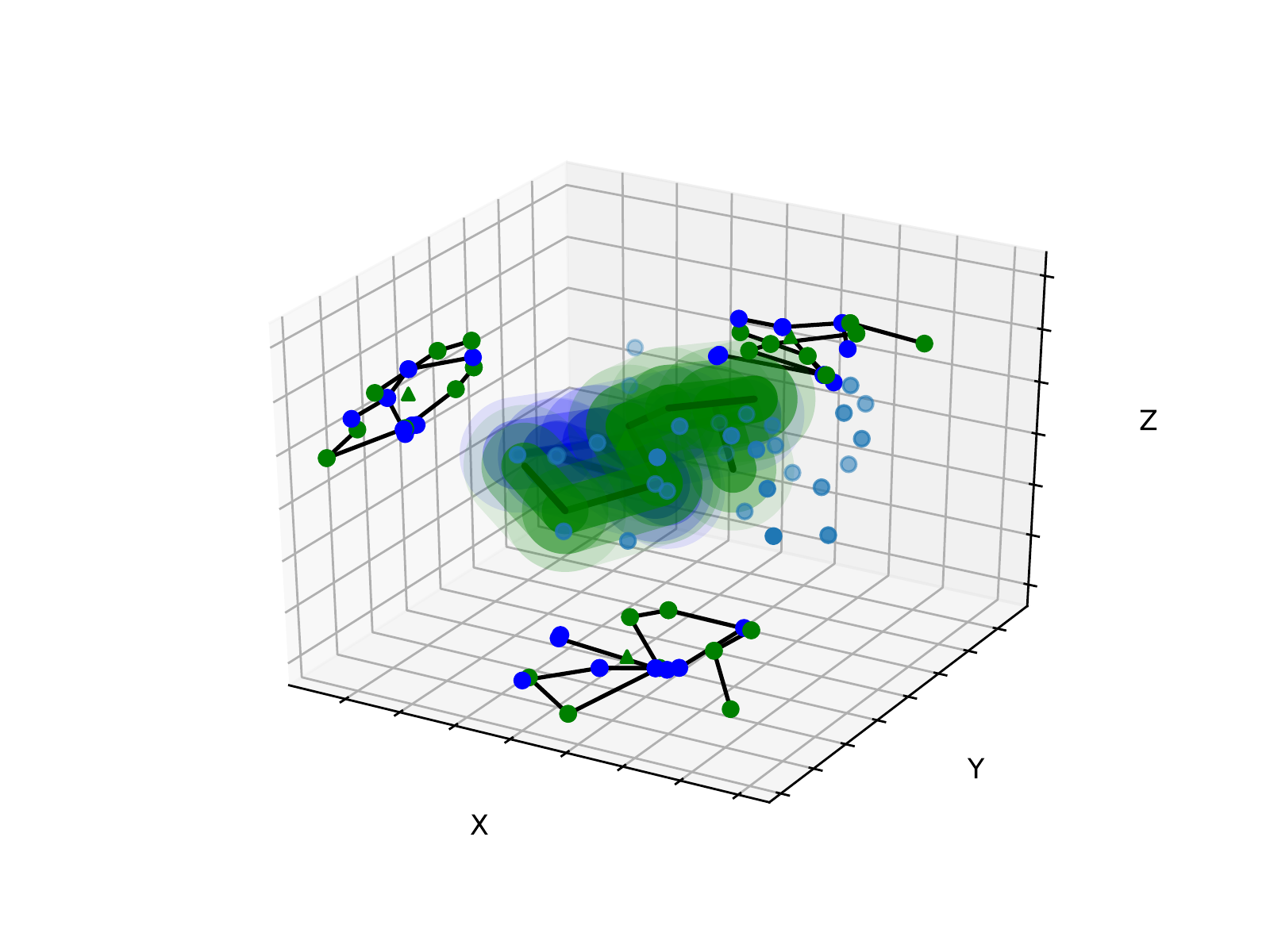}}}
    \\
    \subfloat[Both classifier recognized the toy instance. The object will be assigned by the z-score defined in eq. \ref{zscore} ]{{\includegraphics[width=8cm]{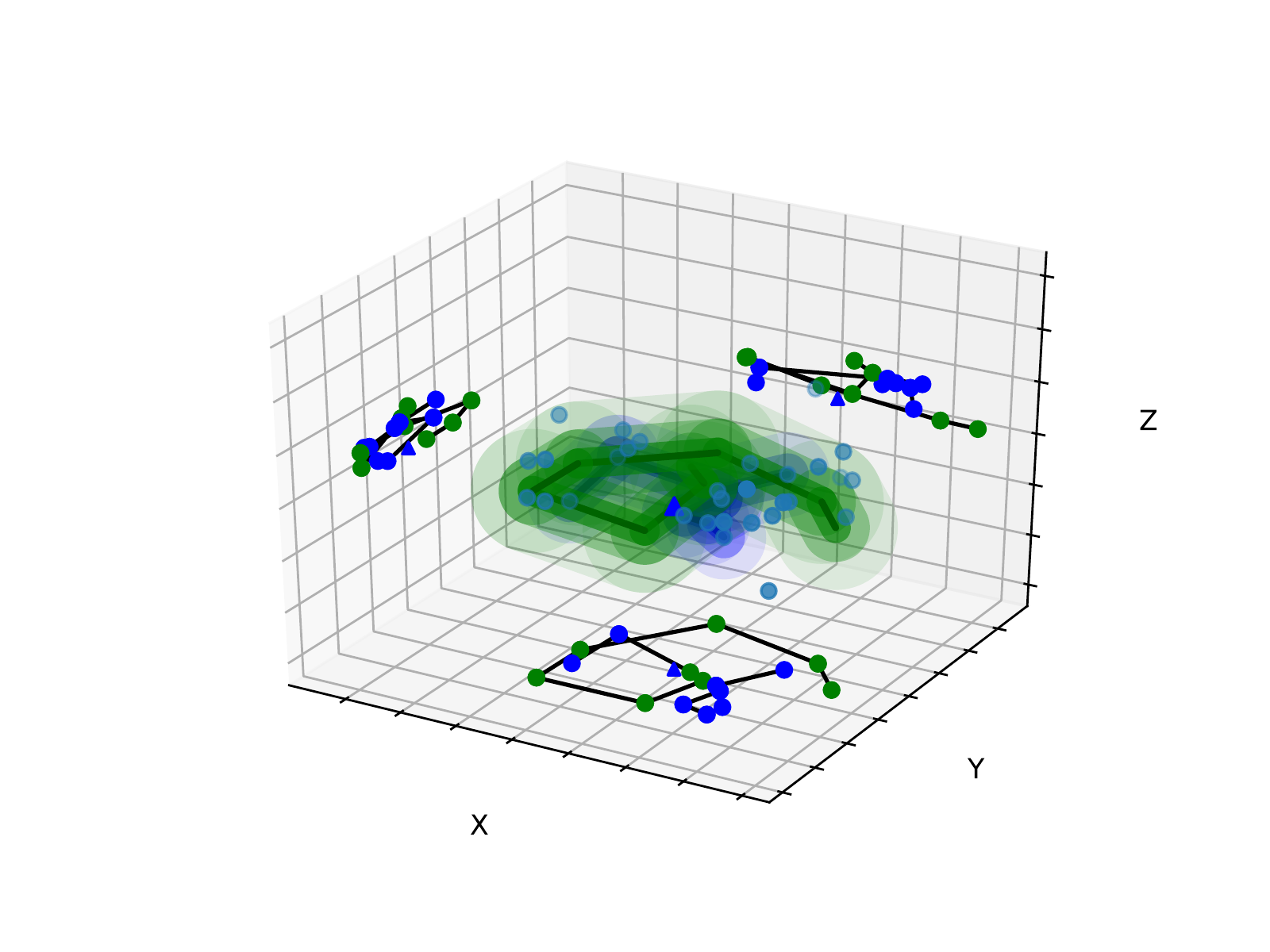}}}
    \caption{In this three figures, we show a 3d toy example of the second step in which two minimum spanning tree per classes are created by neighbours of test instance. In Fig (a) and Fig (b) we show two case where the first/second classifier recognized the test (Triangle blue/green) and the other mst reject. In Fig (c) we observe a simple case which both classifier recognized/rejected the instance. For clarity we plot also the relative orthogonal projection on X, Y, Z.}
    \label{fig:bothmst}
\end{figure}

In the second step, for each test sample $x$ labeled as $w$ (uncertain) in the previous step (see Fig.\ref{fig:main-idea}(b)),  we need to assign one of the two labels: normal (0) or abnormal (1).
For each of such samples we select the $\gamma$ nearest neighbours per classes and create two MST to make the final prediction (see Fig.~\ref{fig:bothmst}).
In this phase we use the label predicted of abnormal class and the ground-truth of the normal class labels.
In~\cite{GrassaGCO19}, we use both structures based on MST, where basic elements of this classifier are not only vertices but also edges of the graph, giving a richer representation of the data. 

Below is a summary of the basic idea of the work presented in~\cite{GrassaGCO19}.
Considering the edges of the graph as target class objects, additional virtual target objects are generated. 
The classification of a new object $x$ is based on the distance to the nearest vertex or edge.
The key of this classifier is to track a shape around the training set not considering only all instances of the training but also edges of the graph, to have more structured information. 
Therefore, in the prediction phase, the classifier considers two important elements:
\begin{itemize}
    \item Projection of point $x$ on an edge defined by vertices ${x_i, x_j}$
    \item Minimum Euclidean distance between $(x, x_i)$ and $(x, x_j)$
\end{itemize}

The Projection of $x$ on an edge $e_{i,j}$ is defined as follow:
\begin{equation}\label{eq:orthogonal_projection}
p_{e_{i,j}}(x)=x_i + \frac{(x_j - x_i)^T(x - x_i)}{||x_j - x_i||^2}(x_j - x_i)
\end{equation}
We check if $p_{e_{i,j}}(x)$ lies on the edge $e_{i,j}$= ($x_i$,$x_j$) then, we compute $p_{e_{i,j}}(x)$ and the Euclidean distance between $x$ and $p_{e_{i,j}}(x)$, more formally if the following condition is true
\begin{equation}\label{eq:orthogonal_projection_check}
0<=\frac{(x_j - x_i)^T(x - x_i)}{||x_j - x_i||^2}<=1
\end{equation}
then
\begin{equation}\label{eq:dist_to_orthogonal_projection}
d(x|e_{i,j})=||x - p_{e_{i,j}}(x)||
\end{equation}
otherwise we compute the Euclidean distance of $x$ and pairs ($x_i$, $x_j$), precisely:
\begin{equation}\label{eq:min_euclidian_dist}
d(x|e_{i,j})=min(||x - x_j||,||x - x_i||)
\end{equation}
Therefore, a new object $x$ is recognized by $MST\_CD\_gp$ (see Algorithm \ref{alg:First step})  if it lies inside the decision boundary that will be described below, otherwise, the object is considered as outlier.
The decision of whether an object is recognized by the classifier or not is based on the threshold $\theta$ of the shape created during the training phase, more formally:
\begin{equation}\label{eq:decision_check}
d_{MST\_CD}(x|X) <= \theta
\end{equation}
where $X$ is the subset of nodes defined by the results obtained in Eqs.~\ref{eq:dist_to_orthogonal_projection} and~\ref{eq:min_euclidian_dist}.

Differently from what is proposed in~\cite{juszczak2009minimum}, where authors set the threshold $\theta$ as a value of the distribution of the edge weights $w_{ij}=||e_{ij}||$ in the given MST, in our approach we enrich this solution with the introduction of additional thresholds.
In~\cite{juszczak2009minimum}, given $\hat{e}=(||e_1||,||e_2||,...,||e_n||)$ as an ordered edge weights values, they define $\theta$ as $\theta = ||e_{[\alpha n]}||$, where $\alpha \in [0,1]$. For instance, with $\alpha=0.5$, we assign the median value of all edge weights into the MST.
In our approach, we set two different thresholds $\theta_0$ and $\theta_1$ to discriminate three different decision boundaries useful to make three kinds of classification. 

In the first step of our approach defined in Algorithm~\ref{alg:First step}, we create only MST, using only the normal class and then, to assign predicted labels to test samples, we introduce the follow discriminative function $f$:

\begin{numcases}{f(\theta_0,\theta_1)=}
    0,                   &  $d_{MST\_CD}(x|X) \leq \theta_0$ \label{dis:0} \\
    \textit{w},    & $\theta_0  < d_{MST\_CD}(x|X) < \theta_1$ \label{dis:1} \\
    1,                   &  $d_{MST\_CD}(x|X) \geq \theta_1$ \label{dis:2}
\end{numcases}

In Eq.~\ref{dis:0} the test instance $x$ relies inside the boundary, therefore the MST assign label $0$ (object recognized) to $x$, otherwise in Eq.~\ref{dis:2} we refuse the object $x$ because is out of the boundary defined by $\theta_1$. 
For each instance $x$ inside the border region defined by the thresholds $\theta_0$ and $\theta_1$, we assign a label $w$ (uncertain object) as defined in Eq.~\ref{dis:1}.
Furthermore, differently from~\cite{juszczak2009minimum}, in our work we do not use all instances of normal class to create a minimum spanning three, but we select $\gamma$ instances from  training set (see graphical representation in Figure~\ref{fig:main-idea}(a)) closest to the test instance $x$ and create a MST.
The main reason we do this is to capture a local representation from an MST built using the neighbors of $x$ from the training set.
After the first step, we obtain an array of predictions as follow:
\begin{equation}\label{eq:prediction}
pred_{y}=(y_{0}, y_{1}, \dots, y_{n})\in \{0, 1, w\}
\end{equation}
where $0$ is the label to represent recognized object as normal and label $1$ represents abnormality.
Labels \textit{w} represent all the test samples inside the border region defined in Eq.\ref{dis:1} of the discriminative function.
In the second step described in Figure~\ref{fig:main-idea}(b), we use a pair of MSTs trained on normal and abnormal class to resolve the ambiguity of all the \textit{uncertain} instances labelled as $w$. 
In our previous work~\cite{GrassaGCO19} in the case in which both classifiers accept/reject test instances, we simply searched the data distribution per classes closest to the test objects and made the final classification. 
In this work, we also extend this function introducing statistical metrics to benefit into classification performance. Given two sets of data samples $N_0$ and $N_1$ containing the $k$ elements per class nearest to the test $x$, we compute standard deviations:
\begin{equation}
\begin{split}
    s_0 = \sqrt{\frac{\sum_{i=1}^{N_{0}} (x_i - \bar{x})^2}{N_0-1} } \quad and \quad
    s_1 = \sqrt{\frac{\sum_{i=1}^{N_{1}} (x_i - \bar{x})^2}{N_1-1} } 
    \end{split}
\end{equation}
where $\textstyle\bar{x}$ is the mean value of these observations.

Finally, given the minimum distance $d$ between a test sample $x$ and an MST node as defined in Algorithm~\ref{alg:second step} rows 9-16 , we define zeta score $\zeta$ as:
\begin{equation}\label{zscore}
\zeta=d * (s + 1)    
\end{equation}
where $s$ is a generic standard deviation of $N$ data samples.
This formulation means that the new observation will be classified using jointly the concept of distance from the appropriate MST and the standard deviation $N$, where we will assign a label class to the test object that obtains the minimum $\zeta$ score. 
More formally we use a function $\hat{f}$ as:
\begin{equation}\label{eq:zeta_score_function}
    \hat{f}(\zeta_0,\zeta_1)= 
\begin{cases}
    0,              & \zeta_0 \leq \zeta_1  \\ 
    1,              & \text{otherwise}
\end{cases}
\end{equation}

\begin{algorithm} \caption{First step}
\label{alg:First step}
\scriptsize
\begin{algorithmic}[1]
\State{$G_0$=\text{All normal instances (train set)}}
\For {$v \in \mathcal G_0 $}
\State all euclidean distances $\gets || x - v ||$
\EndFor
\State NodeX = Take min(all euclidean distances) and return node $v$
\State {mst}, $\theta_{0}, \theta_{1}=$\text{Create small mst(all euclidean distances)}
\State EdgesNodeX $\gets$ Search inc/out edge nodeX and return $(x_i, x_j)$
\For {$x_i,x_j \in EdgesNodeX$ }
\If {$0<=\frac{(x_j - x_i)^T * (x-x_i)}{||x_j - x_i||^2}<=1$}
\State {$P_{e_{_{i_{j}}}}(x)=x_{i} + \frac{(x_{j} - x_{i})^T * (x-x_i)}{||x_j - x_i||^2}*(x_j-x_i)$}
\State {$d(x|e_{_{i_{j}}}) \gets ||x - P_{e_{_{i_{j}}}}(x)||$}
\Else
\State{$d(x|e_{_{i_{j}}}) \gets min \big\{ ||x - x_i||, ||x -x_j||\big\}$}
\EndIf
\EndFor
\State min dist0 = $min(d(x|e_{ij}))$
\State 1 $\gets$  \text{if  $d_{MST\_CD_0}(x|X) \leq \theta_{0}$} 
\State w $\gets$  \text{if  $\theta_{0} < d_{MST\_CD_0}(x|X) < \theta_{1}$} 
\State 0 $\gets$  \text{if  $d_{MST\_CD_0}(x|X) \geq \theta_{1}$}
\end{algorithmic}
\end{algorithm}

\begin{algorithm} \caption{Second Step}
\label{alg:second step}
\scriptsize
\begin{algorithmic}[1]
\State{$G_0$=\text{All normal instances (train set)}}
\State{$G_1$=\text{All abnormal instances predicted in the first step}}
\State{$x$=\text{All uncertain instances from first step}}
\For {$v \in \mathcal G_0 $}
\State all euclidean distances $\gets || x - v ||$
\EndFor
\State NodeX = Take min(all euclidean distances) and return node $v$
\State EdgesNodeX $\gets$ Search inc/out edge nodeX and return $(x_i, x_j)$
\For {$x_i,x_j \in EdgesNodeX$ }
\If {$0<=\frac{(x_j - x_i)^T * (x-x_i)}{||x_j - x_i||^2}<=1$}
\State {$P_{e_{_{i_{j}}}}(x)=x_{i} + \frac{(x_{j} - x_{i})^T * (x-x_i)}{||x_j - x_i||^2}*(x_j-x_i)$}
\State {$d(x|e_{_{i_{j}}}) \gets ||x - P_{e_{_{i_{j}}}}(x)||$}
\Else
\State{$d(x|e_{_{i_{j}}}) \gets min \big\{ ||x - x_i||, ||x -x_j||\big\}$}
\EndIf
\EndFor
\State \textbf{Repeat line 1-16 for graph $G_1$}
\State min dist0 = $min(d(x|e_{ij}))$
\State min dist1 = $min(d1(x|e_{ij}))$
\State Create two mst from neighbours of test x
\State $\theta_0, \theta_1$=Compute a single threshold for both mst
\State 1 $\gets$  \text{if  $d_{MST\_CD_0}(x|X) <= \theta$ and $d_{MST\_CD_1}(x|X) > \theta_1$} \\
0 $\gets$ \text{if  $d_{MST\_CD_0}(x|X) > \theta$ and $d_{MST\_CD_1}(x|X) <= \theta_1$}
\If{\textbf{min dist0 $<= \theta$ and min dist1 $<= \theta_1$}} 
\State $s_{0} = \sqrt{\frac{\sum_{i=1}^{N_{0}} (x_i - \bar{x})^2}{N_0-1}}$
\State $s_{1} = \sqrt{\frac{\sum_{i=1}^{N_{1}} (x_i - \bar{x})^2}{N_1-1}}$
\State $z\_score_{0} = min\ dist_{0} * (s_{0} + 1)$
\State $z\_score_{1} = min\ dist_{1} * (s_{1} + 1)$
\State 0 $\gets$ \text{if $z\_score_{0} < z\_score_{1}$}
\State 1 $\gets$ \text{otherwise}
\EndIf
\If{\textbf{min dist0 $> \theta$ and min dist1 $> \theta_1$}} 
\State \textbf{Repeat line 25-31 for graph $G_1$}
\EndIf
\end{algorithmic}
\end{algorithm}

\begin{algorithm} \scriptsize
\caption{MST CD with gamma parameter}
\begin{algorithmic}[1]
\Function{create small MST}{g0 weight sorted}
\State $small G_0$ = first gamma index sorted values in g0 weight sorted
\State edges couple $\gets$ all combinations nodes small g0
\For {$u,v \in \mathcal edges couple$}
\State $small\ G_0 \gets (node\ u, node\ v, weight=(u,v))$
\EndFor
\State $small\ MST_0 = ComputeMST(small\ G_0)$
\State $e(small\ MST 0) = (||e_0||,||e_1||,..||e_n||)$
\State $\theta_{0}= || e_{(\alpha n )}||$
\State $\theta_{1}= || e_{(\alpha n )}||$\\
\Return \text{small} $MST_0$, $\theta_{0}, \theta_{1}$
\EndFunction
\end{algorithmic}
\label{alg:gamma}
\end{algorithm}

We can summarize the main differences with our previous work as follow:
\begin{enumerate}
    \item we use trained CNNs as deep feature extractors;
    \item we introduce different level of decision boundaries to track uncertain samples and we use strongly rejected instances to create the abnormal class;
    \item we introduce a new discriminative function in case two MSTs accept or reject an instance.
\end{enumerate}


\section{Experimental Results}
\subsection{Datasets}
\begin{figure}
    \centering
    \subfloat{
         \includegraphics[height=1.0cm]{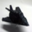}
    }
    \subfloat{
         \includegraphics[height=1.0cm]{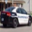}
    }
    \subfloat{
         \includegraphics[height=1.0cm]{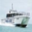}
    }
    \subfloat{
         \includegraphics[height=1.0cm]{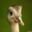}
    }
    \subfloat{
         \includegraphics[height=1.0cm]{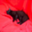}
    }
    \subfloat{
        \includegraphics[height=1.0cm]{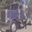}
    }
    \subfloat{
         \includegraphics[height=1.0cm]{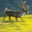}
    }
    \subfloat{
         \includegraphics[height=1.0cm]{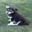}
    }
    \subfloat{
         \includegraphics[height=1.0cm]{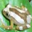}
    }
    \subfloat{
         \includegraphics[height=1.0cm]{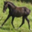}
    }
    
    \subfloat{
         \includegraphics[height=1.0cm]{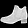}
    }
    \subfloat{
         \includegraphics[height=1.0cm]{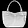}
    }
    \subfloat{
         \includegraphics[height=1.0cm]{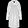}
    }
    \subfloat{
         \includegraphics[height=1.0cm]{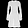}
    }
    \subfloat{
         \includegraphics[height=1.0cm]{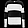}
    }
    \subfloat{
        \includegraphics[height=1.0cm]{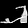}
    }
    \subfloat{
         \includegraphics[height=1.0cm]{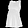}
    }
    \subfloat{
         \includegraphics[height=1.0cm]{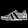}
    }
    \subfloat{
         \includegraphics[height=1.0cm]{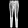}
    }
    \subfloat{
         \includegraphics[height=1.0cm]{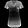}
    }
    \caption{Example of images extracted from CIFAR10 and Fashion MNIST datasets.} 
    \label{fig:datasets}
\end{figure}

To prove the effectiveness of the proposed method we evaluate it on two well-known datasets: Fashion-MNIST \cite{xiao2017fashion} and CIFAR10 \cite{krizhevsky2009learning}. 
Figure~\ref{fig:datasets} shows some examples taken from the two datasets used.
Below, we describe the details on datasets used. 

\textbf{Fashion-MNIST}:
it is a dataset containing 60000 instances for the train set and 10000 instances for the test set. 
The number of classes is 10 and each sample is a 28x28 gray-scale image of clothing. 
This dataset is more challenging than MNIST~\cite{lecun1998mnist} and it represents a useful benchmarker for machine learning algorithms.
Looking at the differences we see that MNIST is too easy. CNNs can reach 99.7\% on MNIST, while classic machine learning algorithms can easily reach 97\%. 
Furthermore, MNIST cannot be representative of modern computer vision problems.

\textbf{Cifar10}:
Consist of 60000 images in 10 classes (6000 per classes) with a training size and test size of 50000 and 10000 respectively.
Each sample is 32x32 color images with a low resolution. The 10 classes are airplanes, cars, birds, cats, deer, dogs, frogs, horses, ships, and trucks. 
Cifar10 is most challenging than Fashion-MNIST due to diverse content and complexity of images. Further, it is widely used as benchmarker comparison by research for classification task.

Although these two datasets are mainly used to study and compare supervised classification techniques, in this work we use them to study a novelty detection algorithm, working with the single classes against all the others.

\subsection{Setup}
Accordingly with novelty detection problems we use only one class of a training set considering all the instances as normal class and performing an $MST\_CD\_gp$ (see Algorithm \ref{alg:First step}) to discover outliers with strong rejection from the test set. 
Then we use the test samples rejected to create the abnormal class and subsequently classify them. 
The test set is composed of 10000 instances, more precisely, 1000 normal samples and 9000 abnormal samples. 
We used the AUC score, according to the literature (\cite{ruff2018deep}) to compare our approach with the others published.
The proposed model was implemented using the framework Pytorch~\cite{paszke2017automatic} and a VGG19~\cite{simonyan2014very} pre-trained on Imagenet~\cite{imagenet_cvpr09} as deep features extractor.
Further, the dimensionality of features extracted is 4096. 
We use these features as input for first step and then for second step of our OCmst model. 
The values of the two thresholds $\theta_0$ and $\theta_1$ were found experimentally using a validation set extracted from the two training data for the two datasets used. 
These parameters have therefore been set as $\theta_0=0.1$ and $\theta_1=0.8$ and have not been modified for all the other experiments.


\subsection{Results}
To better understand the potential of the proposed model, we have conducted two groups of experiments:
\begin{enumerate}
    \item parameter analysis
    \item comparison with other methods proposed in the literature.
\end{enumerate}

For the first group of experiments we extracted a validation set from the training of the CIFAR10 dataset.
We have analyzed the $\gamma$ parameter trying to understand what is the optimal value to use in all the other experiments.
This parameter is very important because the speed and also the memory occupation of the entire novelty detection process depends on it.
For example, on the class Plane of CIFAR10 dataset (10000 test samples), passing from $\gamma = 40$ to $\gamma = 5$ the execution time changes from  $221$ minutes to $74$ minutes.
In Table \ref{tab:gamma variations} we report the AUC score achieved in CIFAR10 using different $\gamma$ values for each experimental run. 
As reported in this table, we choose the best value for the $\gamma$ parameter, i.e. $\gamma = 8$.

In the second group of experiments we compare our approach with many other approaches reported in the literature.
In the Table~\ref{tab:results_cifar} we report results of our method and we compare the same results with the results published in the papers~\cite{scholkopf2001estimating,bishop2006pattern,hadsell2006dimensionality,kingma2013auto,van2016conditional,schlegl2017unsupervised,abati2018and,ruff2018deep,perera2019ocgan} using the CIFAR10 dataset.
From this first group of experiments we can see that our approach is the one that produces the best average results and the best absolute results for 6 classes out of 10 in total.
Another similar experiment is shown in Table~\ref{tab:results_fmnist}. 
In this case we used a different dataset, Fashion-MNIST and we compared ourselves with all the results published in the paper~\cite{schlachterdeep}.
Looking at the average results we can see that our approach ranks fourth but the results are still comparable with the best ones.
We can conclude that the OCmst show the best performance on CIFAR10 and competitive results on Fashion-MNIST.

\begin{table}
\caption{One-class novelty detection results on CIFAR10. Plane and Car are respectively Airplane and Automobile. 
We report the AUC score from different papers (column methods) and then we compare them with our results (last row on the bottom). 
Furthermore, we show the average AUC score for all the One-Class classifiers. 
In bold the best result obtained.
In all the experiments the threshold $\theta_0= 0.8$ is used. 
 }
\label{tab:results_cifar}       
\begin{adjustbox}{max width=\textwidth}
\begin{tabular}{l|ccccccccccc}
\hline\noalign{\smallskip}
Methods & Plane & Car & Bird & Cat & Deer & Dog & Frog & Horse & Ship & Truck & Mean  \\
\noalign{\smallskip}\hline\noalign{\smallskip}
OCSVM \cite{scholkopf2001estimating} & 0.630 & 0.440 & 0.649 & 0.487 & 0.735 & 0.500 & 0.725 & 0.533 & 0.649 & 0.508 & 0.5856 \\
Kde \cite{bishop2006pattern} & 0.658 & 0.520 & 0.657 & 0.497 & 0.727 & 0.496 & 0.758 & 0.564 & 0.680 & 0.540 & 0.6097 \\
Dae \cite{hadsell2006dimensionality} & 0.411 & 0.478 & 0.616 & 0.562 & 0.728 & 0.513 & 0.688 & 0.497 & 0.487 & 0.378 & 0.5358 \\
Vae \cite{kingma2013auto} & 0.700 & 0.386 & 0.679 & 0.535 & 0.748 & 0.523 & 0.687 & 0.493 & 0.696 & 0.386 & 0.5833 \\
Pix CNN \cite{van2016conditional} & 0.788 & 0.428 & 0.617 & 0.574 & 0.511 & 0.571 & 0.422 & 0.454 & 0.715 & 0.426 & 0.5506 \\
Gan \cite{schlegl2017unsupervised} & 0.708 & 0.458 & 0.664 & 0.510 & 0.722 & 0.505 & 0.707 & 0.471 & 0.713 & 0.458 & 0.5916 \\
And \cite{abati2018and} & 0.717 & 0.494 & 0.662 & 0.527 & 0.736 & 0.504 & 0.726 & 0.560 & 0.680 & 0.566 & 0.6172 \\
AnoGan \cite{schlegl2017unsupervised} & 0.671 & 0.547 & 0.529 & 0.545 & 0.651 & 0.603 & 0.585 & 0.625 & 0.758 & 0.665 & 0.6179 \\
Dsvdd \cite{ruff2018deep} & 0.617 & 0.659 & 0.508 & 0.591 & 0.609 & 0.657 & 0.677 & 0.673 & 0.759 & 0.731 & 0.6481 \\
OCGan \cite{perera2019ocgan} & 0.757 & 0.531 & 0.640 & 0.620 & 0.723 & 0.620 & 0.723 & 0.575 & 0.820 & 0.554 & 0.6566 \\  
Soft-Dsvdd \cite{ruff2018deep} & 0.617 & 0.648 & 0.495 & 0.560 & 0.591 & 0.621 & 0.678 & 0.652 & 0.756 & 0.710 & 0.6328 \\
\textbf{OCmst $\gamma=8$}  & 0.742  & \textbf{0.789} & 0.643 & \textbf{0.644} & 0.709  & \textbf{0.688} & \textbf{0.781} & \textbf{0.724} & 0.760 & \textbf{0.817} & \textbf{0.729}  \\
\noalign{\smallskip}\hline
\end{tabular}
\end{adjustbox}
\end{table}

\begin{table}
\caption{Impact of our OCmst varying the gamma parameter using a validation dataset extracted from CIFAR10. 
The row for $\gamma = 8$ shows the best AUC values.}
\label{tab:gamma variations}      
\begin{adjustbox}{max width=\textwidth}
\begin{tabular}{l|cccccccccc}
\hline\noalign{\smallskip}
Methods & Plane & Car & Bird & Cat & Deer & Dog & Frog & Horse & Ship & Truck \\
\noalign{\smallskip}\hline\noalign{\smallskip}
\textbf{OCmst $\gamma=40$}  & 0.691  & 0.755 & 0.632 & 0.630 & 0.671  & 0.653 & 0.744 & 0.711 & 0.721 & 0.754 \\
\textbf{OCmst $\gamma=30$}  & 0.694  & 0.760 & 0.632 & 0.641 & 0.680  & 0.668 & 0.746 & 0.719 & 0.723 & 0.757 \\
\textbf{OCmst $\gamma=20$}  & 0.706  & 0.770 & 0.645 & 0.648 & 0.687  & 0.684 & 0.761 & 0.730 & 0.745 & 0.776 \\
\textbf{OCmst $\gamma=15$}  & 0.722  & 0.766 & 0.642 & 0.626 & 0.683  & 0.655 & 0.731 & 0.722 & 0.726 & 0.786 \\
\textbf{OCmst $\gamma=12$}  & 0.722  & 0.759 & 0.628 & 0.621 & 0.674  & 0.653 & 0.733 & 0.720 & 0.722 & 0.791 \\
\textbf{OCmst $\gamma=8$}  & \textbf{0.781}  & \textbf{0.798} & \textbf{0.665} & \textbf{0.671} & \textbf{0.723}  & \textbf{0.699} & \textbf{0.791} & 0.732 & \textbf{0.777} & \textbf{0.836} \\
\textbf{OCmst $\gamma=5$}  & 0.724  & 0.790 & 0.642 & 0.641 & 0.686  & 0.681 & 0.766 & \textbf{0.736} & 0.753 & 0.809 \\
\noalign{\smallskip}\hline
\end{tabular}
\end{adjustbox}
\end{table}

\begin{table}
\caption{One-class novelty detection results on Fashion-MNIST using AUC score. 
In all the experiments the threshold $\theta_0 = 0.8$.
Three different $\gamma$ values are used and compared with other results published in~\cite{schlachterdeep}}
\label{tab:results_fmnist}       
\begin{adjustbox}{max width=\textwidth}
\begin{tabular}{l|ccccccccccc}
\hline\noalign{\smallskip}
Methods & Ankle & Bag & Coat & Dress & Pullover & Sandal & Shirt & Sneaker & T-shirt & Trouser & Mean  \\
\noalign{\smallskip}\hline\noalign{\smallskip}

OCSVM   &97.8	&79.5	&84.6	&85.9	&85.6	&81.3	&78.6	&97.6	&86.1	&93.9	&87.09\\
IF  &97.9	&88.3	&89.8	&90.1	&87.1	&88.7	&79.7	&98	    &86.8	&97.7	&90.41\\
Imagenet    &78.3	&61.9	&58.3	&60.1	&58.1	&69.2	&57.3	&75.5	&58.1	&75.4	&65.22\\
SSIM    &98.4	&81.6	&87.3	&89.2	&87.2	&85.2	&75.3	&97.8	&83.7	&98.5	&88.42\\
DSVDD   &93.2	&79.1	&87	    &82.9	&83	    &80.3	&74.9	&94.2	&79.1	&94	    &84.77\\
NaiveNN &90.7	&72.9	&80.8	&70	    &73.6	&64	    &71.8	&92	    &62.9	&65.6	&74.43\\
NNwICS  &94.9	&82	    &85.8	&89.1	&82.6	&85.5	&75.6	&94.9	&85.1	&94.6	&87.01\\
Deep OC-ICS &98.5	&88.6	&90.2	&92.1	&88.2	&89.4	&78.3	&98.3	&88.3	&98.9	&\textbf{91.08}\\
\textbf{OCmst $\gamma:$ 15}	&92.77	&85.84	&86.21	&87.52	&77.61	&92.27	&75.47	&94.22	&82.58	&94.12	&86.86\\
\textbf{OCmst $\gamma:$ 10}	&93.04	&86.33	&86.66	&87.85	&78.05	&92.66	&75.58	&94.66	&83	    &94.63	&87.24\\
\textbf{OCmst $\gamma:$ 8}	&93.2	&85.87	&86.8	&87.91	&77.92	&92.69	&75.48	&94.95	&83.15	&94.66	&87.26\\

\noalign{\smallskip}\hline
\end{tabular}
\end{adjustbox}
\end{table}

\section{Conclusion}
In this work we introduce the first hybrid model graph-based for novelty detection problems. 
Our method uses the deep features produced by a convolutional neural network to find a good decision boundary  exploiting  minimum spanning tree structures.
The proposed OCmst outperforms the state-of-the-art in Novelty detection problem on many classes of the CIFAR10 dataset, showing an AUC score higher than others.
In Fashion-MNIST datasets we obtained competitive results.
Our experiments prove the effectiveness of the proposed approach on two different datasets and highlighting advantages and disadvantages.

\begin{acknowledgements}
The authors kindly appreciate the NVIDIA gift of a Titan Xp GPU for this research.
\end{acknowledgements}

\section{Declaration of interest}
The authors declare that they have no conflict of interest.

\bibliographystyle{spmpsci}      
\bibliography{spbasic.bib}
\end{document}